\documentclass[conference]{IEEEtran}
\IEEEoverridecommandlockouts
\usepackage{cite}
\usepackage{amsmath,amssymb,amsfonts}
\usepackage{graphicx}
\usepackage{textcomp}
\usepackage{xcolor}
\usepackage{tabularx}
\usepackage{subcaption}
\usepackage{placeins}
\usepackage{multirow}
\usepackage{color}
\usepackage{soul}
\usepackage{titlesec}
\usepackage[hyphens]{url}
\usepackage{hyperref}
\usepackage[hyphenbreaks]{breakurl}
\usepackage{stfloats}
\usepackage[ruled,vlined]{algorithm2e}
\usepackage{algpseudocode}
\def\BibTeX{{\rm B\kern-.05em{\sc i\kern-.025em b}\kern-.08em
    T\kern-.1667em\lower.7ex\hbox{E}\kern-.125emX}}
\usepackage{array}
\newcolumntype{P}[1]{>{\centering\arraybackslash}p{#1}}
\newcolumntype{M}[1]{>{\centering\arraybackslash}m{#1}}

\makeatletter
\newcommand*\bigcdot{\mathpalette\bigcdot@{.5}}
\newcommand*\bigcdot@[2]{\mathbin{\vcenter{\hbox{\scalebox{#2}{$\m@th#1\bullet$}}}}}
\makeatother

\makeatletter
\def\ps@IEEEtitlepagestyle{
  \def\@oddfoot{\mycopyrightnotice}
  \def\@evenfoot{}
}
\def\mycopyrightnotice{
  {\footnotesize 979-8-3503-8713-1/24/\$31.00~\copyright~2024 IEEE\hfill}
  \gdef\mycopyrightnotice{}
}

\makeatother
\usepackage{eso-pic}
\newcommand\AtPageUpperMyright[1]{\AtPageUpperLeft{
 \put(\LenToUnit{0.5\paperwidth},\LenToUnit{-1cm}){
     \parbox{0.5\textwidth}{\raggedleft\fontsize{9}{11}\selectfont #1}}
 }}
\newcommand{\conf}[1]{
\AddToShipoutPictureBG*{
\AtPageUpperMyright{#1}
}
}

\conf{IEEE International Parallel and Distributed Processing Symposium (IPDPS) June $3^{\text{rd}}$ - $7^{\text{th}}$, 2025}

\begin{document}

\title{Longer Attention Span: Increasing Transformer Context Length with Sparse Graph Processing Techniques}

\author{\IEEEauthorblockN{Nathaniel Tomczak}
\IEEEauthorblockA{\textit{Computer and Data Sciences} \\
\textit{Case Western Reserve University}\\
Cleveland, OH, U.S.A. \\
nkt8@case.edu}
\and
% \IEEEauthorblockN{Manish Routhu}
% \IEEEauthorblockA{\textit{Computer and Data Sciences} \\
% \textit{Case Western Reserve University}\\
% Cleveland, OH, U.S.A. \\
% mxr809@case.edu}\\
% \and
\IEEEauthorblockN{Sanmukh Kuppannagari}
\IEEEauthorblockA{\textit{Computer and Data Sciences} \\
\textit{Case Western Reserve University}\\
Cleveland, OH, U.S.A. \\
sxk1942@case.edu}}
% \and
% \IEEEauthorblockN{Sai Dheeraj Yanduru}
% \IEEEauthorblockA{\textit{Computer and Data Sciences} \\
% \textit{Case Western Reserve University}\\
% Cleveland, OH, U.S.A. \\
% sxy874@case.edu}
% }
% \and
% \and

\maketitle

\begin{abstract}
Transformers have demonstrated great success in numerous domains including natural language processing and bioinformatics. This success stems from the use of the attention mechanism by these models in order to represent and propagate pairwise interactions between individual tokens of sequential data. However, the primary limitation of this operation is its quadratic memory and time complexity in relation to the input’s context length – the length of a sequence over which the interactions need to be captured. This significantly limits the length of sequences that can be inferred upon by these models. Extensive research has been conducted to reduce the number of pairwise interactions to sub-quadratic in relation to the context length by introducing sparsity into the attention mechanism through the development of sparse attention masks. However, efficient implementations that achieve “true sparsity” are lacking. 

In this work, we address this issue by proposing a graph computing view of attention where tokens are perceived as nodes of the graph and the attention mask determines the edges of the graph. Using this view, we develop graph processing algorithms to implement the attention mechanism. Both theoretically and empirically, we demonstrate that our algorithms only perform the needed computations, i.e., they are work optimal. We also perform extensive experimentation using popular attention masks to explore the impact of sparsity on execution time and achievable context length. Our experiments demonstrate significant speedups in execution times compared to state-of-the-art attention implementations such as FlashAttention for large sequence lengths. We also demonstrate that our algorithms are able to achieve extremely long sequence lengths of as high as 160 million on a single NVIDIA A100 GPU (SXM4 80GB).

GitHub: \url{https://github.com/KLab-AI3/Graph-Processing-Attention-IPDPS-2025}
\end{abstract}


% \vspace*{-2.5cm}

\section{Introduction}

%We should keep this paper pretty focused. Our objective is to exploit sparsity on attention to obtain the following: (a) better runtime for a given sequence length, (b) achieving longer sequence length. Our claim is that existing techniques do not achieve this sparsity and so we propose a graph view to exploit it. 

% Para - 1: Talk about importance of Transformer models in various applications. 

Transformers have emerged as one of the most successful AI/ML techniques for modeling sequence data. They have drastically impacted a variety of scientific and engineering applications such as language modeling~\cite{zhao2024surveylargelanguagemodels}, molecular design~\cite{bhowmik2024enhancing}, cancer pathology classification~\cite{chandrashekar2024path}, genomic sequence modeling~\cite{nguyen2024hyenadna}, and others. These models derive their power from an operation called the \textit{attention mechanism}. Given a sequence of tokens of length $L$, the attention mechanism extracts pairwise similarities between the tokens~\cite{vaswani2017attention}.

%Para - 2: Talk about attention mechanism, and how it is the limiting factor in computations of Transformer models.
However, the time and memory complexity of the attention mechanism is quadratic in $L$ as it captures interactions between each pair of the sequence. This has severely limited the context length, i.e., the length of a sequence over which a transformer model can capture these interactions (also know as sequence length in the literature). For applications such as genomics, at least 4-5 orders of magnitude of increase in context length is needed~\cite{nguyen2024hyenadna}.

%Para - 3: Talk about sparse attention mechanisms and how they are proposed to reduce the computational complexity
A popular approach for reducing the computational complexity is to introduce sparsity in the attention mechanism~\cite{zaheer2020big,ding2023longnet,beltagy2020longformer}. In this approach, an $L \times L$ 0-1 attention mask is used to capture the interactions between pairs. The mask is made sparse by reducing the number of interacting pairs (making the corresponding entries 0) in a structured manner. Local, dilated windowed, block, and random are popular patterns that are used to introduce sparsity while ensuring that critical interactions are preserved to maintain the accuracy of the models. \cite{ding2023longnet} has demonstrated that, even with an exponentially decreasing number of connections with respect to the distance between pairs, the test perplexity scores remain comparable to dense models. This demonstrates that utilizing sparsity to achieve ultra-long sequence modeling is a feasible approach.

%para - 4: However, these techniques are not "hardware aware". Existing implementations still rely on dense dense, followed by masking. This requires O(L^2) space and O(L^3) time. 
However, existing implementations of sparse attention mechanisms are not ``hardware-aware". They perform dense-dense matrix multiplication followed by invalidation of entries based on the attention mask \cite{pytorch-sdp, xFormers2022}. Optimizations have focused on partitioning the matrices into blocks, applying matrix reordering techniques to obtain denser blocks, and performing block matrix multiplication only on the blocks that have non-zero entries in the mask (see Section \ref{sec:rel_works} for details). However, these techniques end up performing excess computations as the block may contain several 0 mask elements.   

In this work, we aim to achieve ``\textbf{true sparsity}", meaning that only the necessary computations are performed for any arbitrary attention mask. We accomplish this by posing attention as a graph problem with individual tokens of the sequence perceived as nodes and the interactions, as defined by the attention mask, determining the edges between the nodes. With this perspective, we develop efficient algorithms for masked attention with various mask patterns. The contributions of this paper are:
\begin{itemize}
    \item We propose a graph computing perspective of the attention operation, enabling us to realize a reduction in computations due to sparse attention mechanisms.
    \item Using this view, we develop graph algorithms that are work-optimal over arbitrary attention masks.
    \item For local, global, and dilated attention, we further optimize our graph algorithms to achieve higher performance through longer context lengths.
   % \item We implement a PyTorch back-end using our implementations to seamlessly integrate our algorithms into existing LLMs using PyTorch.
    \item We conduct thorough experiments to demonstrate the benefits of our graph computing perspective where we obtain a 4.46x and 51.06x speedup relative to FlashAttention for sequence lengths of 2,097,152 and 160 million, respectively. Our implementations are able to achieve these context lengths on a single NVIDIA A100 GPU (80 GB), paving the way for training models that can handle extremely long sequences using distributed training. 
    \item We create a PyTorch back-end for our algorithms for seamless integration into existing LLMs.
\end{itemize}

%para - 5: The objective of this work is to actually realize the benefits of sparse attention masks. To achieve this, we model the attention, which by definition is defined as capturing pairwise interactions, as a graph computation as opposed to the existing tensor based computations. Using our implementation, we achieve a speedup of xx and increase in sequence length on a single GPU node of xx against existing implementations. The contributions of this paper are:
% \begin{itemize}
%     \item We propose a graph computing view of attention mechanism. This view enables us to realize reduction in computations due to sparse attention mechanisms.
%\item Using this view, we develop two algorithms using both CSR and COO representations for the sparse attention mask.
%\item We implement a pytorch backend using our implementations to seamlessly integrate our algorithms into existing LLMs using pytorch.
%\item we conduct thorough experimental results ...
% \end{itemize}

% \vspace*{-2.0cm}

\section{Background}

\subsection{Transformer Models}
\label{ssec:tx}
%A high level description of Transformer models and the various constituent layers. 

Transformer models are encoder-decoder architectures consisting of a sequence of ``transformer layers". A transformer layer takes, as input, a sequence of tokens of length $L$, where each token is a $d$ dimensional vector. This sequence is projected into a set of $d_k$ dimensional queries packed in a matrix $Q \in R^{L \times d_k}$, a set of $d_k$ dimensional keys packed in a matrix $K\in R^{L \times d_k}$, and a set of $d_v$ dimensional values, packed into a matrix $V\in R^{L \times d_v}$ using learnable weight matrices $W_Q, W_K$ and $W_v$ respectively \cite{vaswani2017attention}. This operation can be represented using the following equation: 

\vspace{-0.5cm}

\begin{small}
\begin{flalign}
    Attention(K,Q,V) = softmax(\frac{QK^T}{\sqrt{d_k}})V \label{eqn:attn}
\end{flalign}
\end{small}

\vspace{-0.2cm}

In practice, a multi-headed attention mechanism is utilized where $Q$, $K$, and $V$ are sliced into several smaller dimensions, attention is applied in parallel to each projection, and the results are concatenated to obtain the attention output~\cite{vaswani2017attention}. Following the attention operation, a learnable fully connected network is used to project the output into a $L \times d$ dimension matrix to be processed by the subsequent transformer layer.

%Transformer models are encoder-decoder architectures where each transformer layer takes as input a sequence of $d$ dimensional tokens of length $L$, projects them into a set of $d_k$ dimensional queries packed in a matrix $Q \in R^{L \times d_k}$, a set of $d_k$ dimensional keys packed in a matrix $K\in R^{L \times d_k}$, and a set of $d_v$ dimensional values, packed into a matrix $V\in R^{L \times d_v}$ using learnable weight matrices $W_Q, W_K$ and $W_v$ respectively \cite{vaswani2017attention}.

%Figure \ref{eqn:attn} demonstrates the computations performed in the attention process. The three matrices from left to right represent the queries, keys, and values corresponding to the $L$ input tokens that are obtained by projection using the learnable weight matrices. The query matrix (leftmost matrix) stores the $d_k$ dimensional queries as rows, the key matrix (middle matrix) stores the $d_k$ dimensional keys as columns, and the value matrix (rightmost matrix) stores the $d_v$ dimensional values as rows. Mathematically, the attention computation can be represented using Equation \ref{eqn:attn}:

% \begin{flalign}
%     Attention(K,Q,V) = softmax(\frac{QK^T}{\sqrt{d_k}})V \label{eqn:attn}
% \end{flalign}

% In practice, a multi-headed attention mechanism is utilized, where the keys, queries, and values are projected to several smaller dimensions, attention is applied in parallel to each projection, and the results are concatenated to obtain the attention output~\cite{vaswani2017attention}.

\subsection{Attention Mechanism}
\label{ssec:attn}
%Describe attention mechanism along with the equation. We can also write the intuitive definition similar to the hpec paper here. 

\begin{figure}
    \centering
    \includegraphics[width=0.7\linewidth]{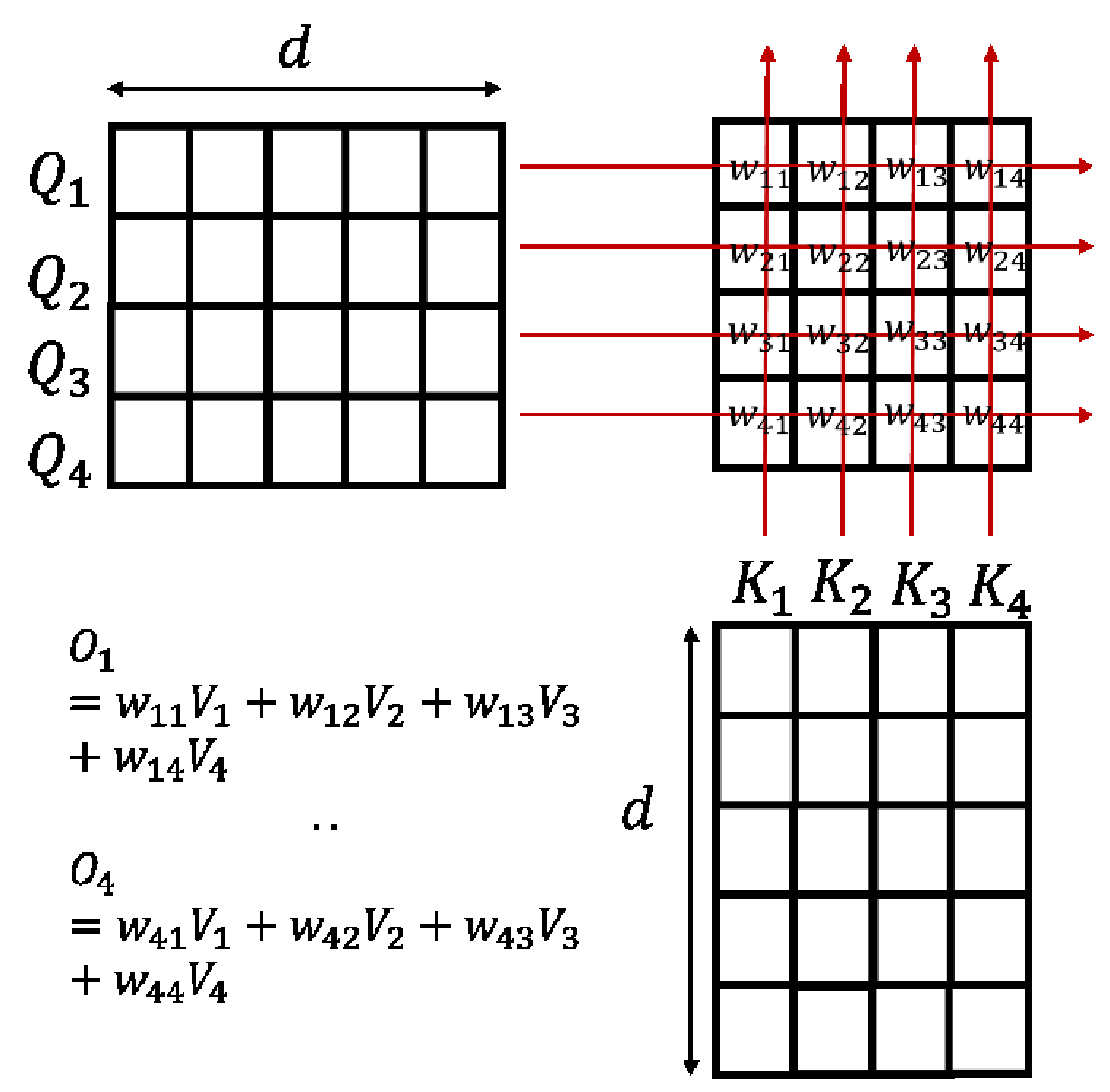}
    \caption{Visualization of the attention mechanism. $w_{ij}$ is the similarity of Query $i$ and Key $j$. The similarities are used to weight the Values.}
    \label{fig:qk-graph}
\end{figure}

Figure~\ref{fig:qk-graph} provides a graphical perspective of the attention mechanism. When viewed as matrices, attention performs matrix multiplication between the query and key matrices, row-wise softmax on the output, and another matrix multiplication between the softmax result and the value matrix. 

Intuitively, attention performs a soft lookup of values for each query. Specifically, for a query $i$ (a row of the $Q$ matrix in Figure~\ref{fig:qk-graph}), its dot product with columns of the $K$ matrix followed by a softmax on the output produces a probability vector (i.e., all entries are non-negative and sum up to one). A linear combination of value vectors (rows of the matrix $V$) with the weights of the probability vector produces the final output corresponding to the query $i$. \textit{In other words, instead of outputting a single value vector as performed in a hard lookup, a combination of values, dependent upon similarities between the queries and the keys is used to produce the output vector.}

\subsection{Sparsity in Attention}
\label{ssec:sparsity}
%Describe how various works obtain sparsity in attention. Talk about various types of attention patterns.
This soft lookup perspective of attention motivated the introduction of sparsity into the mechanism. Researchers have identified that, for each query, only the most important terms, as determined by the similarity matrix resulting from key-query interactions, need to be considered~\cite{kitaev2020reformer}. Furthermore, to determine similar keys, without computing the key-query dot products, researchers have proposed various connectivity patterns between keys and queries to capture their similarity~\cite{zaheer2020big,ding2023longnet,beltagy2020longformer}. These patterns are represented in $L \times L$ 0-1 attention masks where the rows denote queries and the columns denote keys. As both queries and keys are projections of the same $L$ sized sequence, the attention mask can be perceived as capturing important pairwise similarities between the tokens of the sequence.

\begin{figure}[!ht]
    \centering
    \includegraphics[width=0.45\textwidth]{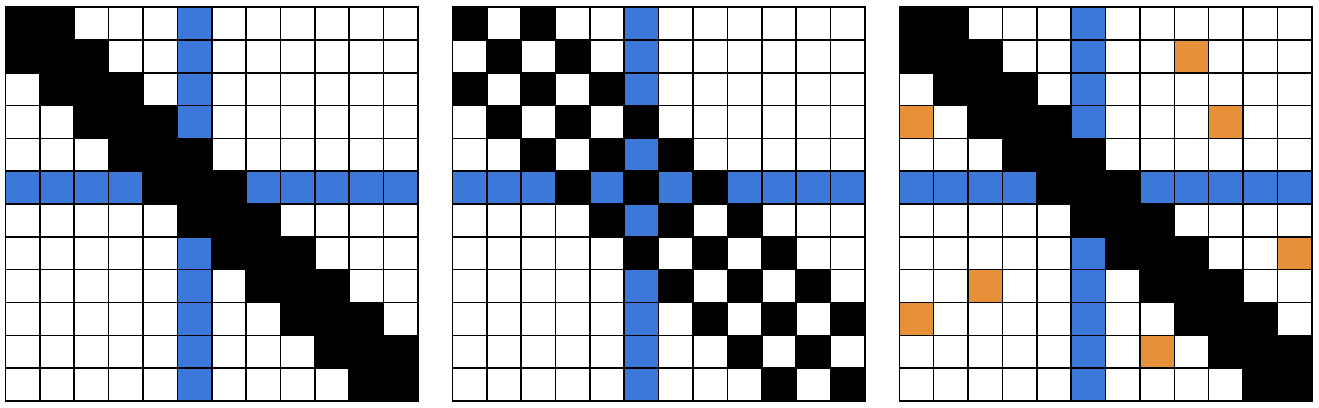}
    \vspace{-0.15cm}
    \caption{Visualizing different masks from the Longformer (left and center) and BigBird (right) transformers. The white cells indicate masked terms that are not considered. The black cells are local attention, which can be dilated. The blue cells correspond to global attention. The orange cells are uniform random attention.}
    \label{fig:ex_4_masks}
\end{figure}

Figure \ref{fig:ex_4_masks} shows common attention patterns and their combinations used by two transformers: the left and central masks are utilized by Longformer and the right by BigBird \cite{beltagy2020longformer, zaheer2020big}. The white cells correspond to terms that are masked from the calculation. The forms of attention in relation to Figure \ref{fig:ex_4_masks} are described below.

\noindent \textbf{Local/Windowed Attention:} The black cells represent local attention which gives a token the ability to look $n$ tokens forwards and backwards from itself \cite{beltagy2020longformer, zaheer2020big, child2019generating}.

\noindent \textbf{1D Dilated Windowed Attention:} Local attention can also be dilated as shown in the central figure. This creates uniform gaps of a set size between each attended token that widens the effective view distance for each token \cite{beltagy2020longformer}. For an index $(i,j)$, a window size of $w$, and a dilation factor $r$, we can determine if a value is masked (returns 0) or not (returns 1) by:

\begin{scriptsize}
\begin{verbatim}
if ((abs(i - j) < w) && (abs(i - j) % (r + 1) == 0)) {
  return 1;
} else {
  return 0;
}
\end{verbatim}
\end{scriptsize}

\noindent \textbf{2D Dilated Windowed Attention:} Dilation can also occur over two dimensions with blocks \cite{ding2023longnet}. Given a context length $L$, an index $(i,j)$, a block height and width of $b$, and a dilation factor $r$, we can determine if a value is masked by:

\begin{scriptsize}
\begin{verbatim}
if (floor(i/(L/b)) == floor(j/(L/b))) {
  i_b = i % b;
  j_b = j % b;
  if ((i_b % (r + 1) == 0) && (j_b % (r + 1) == 0)) {
    return 1;
  } else {
    return 0;
  }
} else {
  return 0;
}
\end{verbatim}
\end{scriptsize}

\noindent \textbf{Global Attention:} The blue is global attention and is reserved for designated tokens that can attend to all other tokens in the sequence \cite{zaheer2020big, wu2021representing}.

\noindent \textbf{Random Attention:} The orange indicates token-token relationships that are chosen from a uniform random distribution \cite{zaheer2020big}.

These and other forms of attention are often represented by blocks larger than 1 token, called a block sparse format \cite{zaheer2020big, liu2023ring, liu2024blockwise}. This representation allows for good utilization of GPU resources, however it restricts the resolution of sparsity that can be achieved. The work in this paper removes resolution restrictions on imposed sparsity without requiring excess masking computations. 

\subsection{How Sparse are the Masks?}
\label{ssec:how_sparse}

A natural question to ask at this point is: What are the typical sparsity levels of these masks? As, depending upon the level, a graph computing view may or may not be justified. We first define a term known as the sparsity factor, $S_f$, which can be calculated by the equation:

\vspace{-0.2cm}

\begin{small}
\begin{equation}
    S_f = \frac{NNZ}{TE}
\end{equation}
\end{small}

\vspace{-0.1cm}

\noindent where $NNZ$ is the number of non-zero elements in the mask and $TE$ is the total number of elements in the mask. $S_f$ is constant for a mask in the range $[0, 1]$ where zero is a fully sparse matrix and one is a fully dense matrix. It operates as a multiplicative constant for the number of dot products required to build the attention matrix: $\mathcal{O}(S_fL^2)$. 

Popular attention masks from Bigbird \cite{zaheer2020big}, and Longformer \cite{beltagy2020longformer} achieve $\mathcal{O}(L)$ dot products. In other words, the sparsity factor $S_f$ reduces as $\mathcal{O}(\frac{1}{L})$ with context length. However, the hidden constant term is not clear to obtain a concrete sparsity value.

To obtain this number, we refer to LongNet \cite{ding2023longnet} which shows that the number of dot products needed is $\frac{2\alpha}{(\alpha-1)}w_oL$ where $\alpha$ is the geometric series ratio used to compute the segment length and dilation parameters in their algorithm and $w_0$ is the first element of the segment length geometric series. Setting $\alpha = 2$ and $w_o = 2048$ based on the parameters they use in the experimental results, this number equals to $2730L$ leading to a sparsity factor of $\frac{2730}{L}$. The value of sparsity factor for selected sequence lengths of $\{16\text{k}, 32\text{k}, 1 \text{ million}, ...,160 \text{ million}, 1 \text{ billion} \}$ will be $\{0.17, 0.085, 0.0027,...,0.000017,2.7e^{-6}\}$. This implies that, for large sequence lengths, a graph computing view, as proposed in this work, is very well motivated and has the potential to significantly improve upon the existing tensor based implementations.

%\subsection{Graph Computations}

%Talk about computations on graph, when it should be used. we can also mention graphBLAS here. 

\section{Related Works}
\label{sec:rel_works}

As the key objective of this work is to increase context length by exploiting sparsity, we discuss relevant works that focus on either achieving efficient sequence parallelism or on achieving efficient implementation of sparse masks.

\noindent \textbf{Sequence Parallelism:} DeepSpeed Ulysses~\cite{jacobs2023deepspeed} achieves sequence parallelism for the original $L^2$ complexity attention by distributing the partitions of the $K,Q,V$ matrices along the sequence dimensions across computing nodes. Megatron~\cite{shoeybi2019megatron} performs sequence parallelism that is tightly coupled with their tensor parallelism~\cite{jacobs2023deepspeed}. Ring attention achieves sequence parallelism for block sparse attention masks~\cite{liu2023ring}. LongNet achieves sequence parallelism using a dilated attention mask and requires all-gather of $K,Q$ matrices~\cite{ding2023longnet}. On a single node, these techniques still rely on dense matrix operations for attention calculations. Our technique, that exploits sparsity, is orthogonal to these methods and has the potential to significantly scale the sequence length that can be achieved using these techniques for a fixed number of nodes. 

\noindent \textbf{Efficient Sparse Attention Implementations:} While works such as BigBird \cite{zaheer2020big}, Longformer \cite{beltagy2020longformer}, and Reformer \cite{kitaev2020reformer} proposed sparse attention masks, their implementations in libraries such as PyTorch \cite{pytorch-sdp} and xFormers \cite{xFormers2022} still rely on dense operations. Specifically, they employ a variant of the Scaled Dot Product (SDP) attention \cite{vaswani2017attention} where they first perform a dense matrix multiplication of $Q$ and $K$ that can incorporate block sparsity for low-resolution masking, set the excess terms corresponding to the zero entries in the attention mask to $-\infty$, perform a row-wise softmax (which results in $-\infty$ getting converted to 0) and finally a sparse-dense matrix multiplication is performed between the resultant matrix (sparse) and the $V$ matrix (dense).   

As a result, FlashAttention --- a highly optimized version of SDP --- has remained the most efficient attention implementation despite the fact that it performs the full $L^2$ computations \cite{dao2022flashattentionfastmemoryefficientexact, dao2023flashattention2fasterattentionbetter, shah2024flashattention3fastaccurateattention}. Recent works \cite{pagliardini2023fastercausalattentionlarge}, \cite{sharma2024efficientlydispatchingflashattention} have enabled support for sparsity by partitioning the attention mask and the $K,Q,V$ matrices into blocks and only computing the blocks that have at least one non-zero element using the FlashAttention algorithm.

\cite{pagliardini2023fastercausalattentionlarge}, \cite{sharma2024efficientlydispatchingflashattention}, while improving upon FlashAttention in exploiting sparsity, do not achieve true sparsity. Specifically, they still compute dot products corresponding to the 0 entries in the blocks leading to $O(d)$ unnecessary computations, where $d$ is the embedding dimension. In contrast, our work achieves true sparsity and only performs the computations that are needed.   

%Subsection 1 - We should talk about works that focus on increasing the sequence length - look at my proposal. The conclusion here could be that they do not exploit sparsity. They simply distribute across nodes to achieve larger sequence length.

% Subsection 2 - works on data, model/pipeline parallelism - deepspeed, megatron, etc. Here again, we can mention that they typically use dense attention computations and the parallelism they achieve is totally orthogonal. 

% Kernel Methods - We can briefly mention them, but they are not relevant here.

% conclusion - Our conclusion here is that we are reducing the computations by exploiting sparsity which none of the above works do. Our work is orthogonal to these other works and can be integrated with them to achieve much higher scalability. 

%define \textbf{true sparsity} in the introduction, but we can bring it up again here and say many methods approximate sparsity (often using some form of block-sparsity). Nothing reaches "truly sparse" levels of work... i.e. nothing is wasted

% https://arxiv.org/abs/2306.01160 --- Similar to the one below, but published before

% https://arxiv.org/abs/2409.15097 --- this is the new paper that dynamically creates a block-sparse representation of an arbitrary graph

\section{Graph Computing View of Attention}

\subsection{Modeling}
Our graph computing view of attention is motivated by its intuitive definition described in Section~\ref{ssec:attn}. Specifically, for each token $i$, the corresponding outputs are determined by its query, and the keys and values of a subset of tokens to which it is closely related to. Thus, instead of modeling it as tensor operations, as performed by state-of-the-art deep learning frameworks, we describe attention as a graph computation below.

For an input $d_k$ dimensional token sequence $X_1, X_2,\dots,X_L$, we build a graph $G$ with vertex set $V_G = v_1, v_2, \dots, v_L$. The attributes of the vertices are the projected keys, queries, and values. The edge set $E_G$ is determined by the $L \times L$ attention mask ($A$). Specifically, a directional edge exists between vertex $v_i$ and $v_j$ if the $A_{ij} = 1$. 

%In this work, we solely focus on developing graph algorithms for the attention mechanism. However, it 

%with attributes as the corresponding $d$ dimensional tokens and no edges. The projection matrices $W_Q, W_K, W_V$ are applied to each vertex separately to produce three vector attributes for each vertex. Now, when we need to compute the attention mechanism, an edge set $E$ is overlaid onto the graph. The edge set represents the attention mask and may vary for different layers. 

\subsection{Graph Algorithms for Attention}
We develop two algorithms based on the graph model of attention. Both algorithms are single-batch and single-headed to facilitate focus on the experiments, though it is trivial to scale them to a multi-headed approach. Both algorithms were inspired by FlashAttention and utilize the online softmax operation \cite{milakov2018onlinenormalizercalculationsoftmax, dao2022flashattentionfastmemoryefficientexact}. The first uses an explicit input mask while the second has an implicit mask that is calculated from input parameters as the kernel executes. The six graph processing algorithms that were implemented are as follows, subset by their mask type:

\begin{itemize}
    \item \textbf{Explicit Mask}
    \begin{itemize}
        \item \textbf{COO (coordinate):} The row indices, column indices, and values vectors are passed in.
        \item \textbf{CSR (compressed sparse row):} The row offset, column indices, and values vectors are passed in.
    \end{itemize}
    \item \textbf{Implicit Mask}
    \begin{itemize}
        \item \textbf{Local:} A window size, $n$, is passed in and the window indices are calculated relative to the index token of a row.
        \item \textbf{1D Dilation:} A window size, $w$, and a dilation factor, $r$, are passed in and the mask indices are calculated relative to the index token of a row.
        \item \textbf{2D Dilation:} A block height and width, $b$, and dilation factor, $r$, are passed in and the mask indices are calculated relative to the index token of a row.
        \item \textbf{Global (non-local):} A vector of global token indices and window size, $n$, are passed in. Attention indices are calculated for both the global and local mask and then the local mask is subtracted from the global.
    \end{itemize}
\end{itemize}

The COO and CSR versions for the first algorithm can handle any arbitrary attention mask and their performance is compared through microbenchmarks in Section \ref{subsec:microbench}. The local, 1D dilation, 2D dilation, and global (for simplicity, we will drop \textit{non-local} for the rest of the paper) implementations are inflexible attention masks outside of their parameter inputs, we will refer to these collectively as ``\textbf{ordered sparsity}" patterns within the paper. Both algorithmic approaches can be described by Algorithm \ref{alg:gpatt}. 

\begin{algorithm}
    \caption{Graph Processing Attention}\label{alg:gpatt}
    \KwIn{Attention-specific parameters $P_a$ and}
    $\;\;\;\;\;\;\;\;\;\;$ Matrices $Q, K, V \; \in \mathbb{R}^{L \times d_k}$ and \\
    $\;\;\;\;\;\;\;\;\;\;$ Graph $G = (V_G, E_G)$ with vertex attributes \\
    $\;\;\;\;\;\;\;\;\;\;\; V_G = \{v \in \mathbb{R}^L : \forall \; v_i \in  v \; \exists \; Q_i, K_i, V_i \in \mathbb{R}^{d_k}$ \\
    $\;\;\;\;\;\;\;\;\;\;\;\;\;\;\;\;\;\;\;\;\;\;\;\;\;\;\;\;\;\;\;\;\;\;\;\; \text{such that} \; v_i = (Q_i, K_i, V_i)\}$ \\
    \KwOut{$O \in \mathbb{R}^{L \times d_k}$}
    % \BlankLine
    \hrulefill
    \BlankLine
    \textbf{Initialize:} $O \in \mathbb{R}^{L \times d_k} \gets 0 \; \forall \; o_{i,j} \in O$, \\
    $\;\;\;\;\;\;\;\;\;\;\;\;\;\;\;\;\;\; l \in \mathbb{R}^L \gets 0 \; \forall \; l_i \in l$, \\
    $\;\;\;\;\;\;\;\;\;\;\;\;\;\;\;\; m \in \mathbb{R}^L \gets - \infty \; \forall \; m_i \in m$ \\
    \For(\textbf{in parallel}){$1 \leq i \leq L$}{
        \texttt{vi\_Neighbors} $\gets \texttt{\textbf{Get\_Neighbors}}(G, i, P_a)$ \\
        \For{$j \in \texttt{vi\_Neighbors}$}{
        $K \gets \texttt{\textbf{Pull}}(K_j)$ \\
        $W \gets Q_i \bigcdot K$ \\
        $m_i^{new} \gets \texttt{\textbf{max}}(m_i, W)$ \\
        $l_i^{new} \gets l_i * \texttt{\textbf{exp}}(m_i - m_i^{new}) + \texttt{\textbf{exp}}(W - m_i^{new})$ \\
        $V \gets \texttt{\textbf{Pull}}(V_j)$ \\
        $O_i \gets (l_i^{new})^{-1} * \big[ l_i * \texttt{\textbf{exp}}(m_i - m_i^{new}) * O_i \; +$ \\
        $\;\;\;\;\;\;\;\;\;\;\;\;\;\;\;\;\;\;\;\;\;\;\;\;\;\;\;\; \texttt{\textbf{exp}}(W - m_i^{new}) * V \big]$ \\
        $l_i \gets l_i^{new}, m_i \gets m_i^{new}$
        }
    }
    \textbf{Return:} $O$
\end{algorithm}

All implementations utilize a shared memory model coupled with online softmax \cite{milakov2018onlinenormalizercalculationsoftmax, dao2022flashattentionfastmemoryefficientexact} and are parallelized along the $L$ dimension, simultaneously operating on rows of the attention matrix. The primary difference lies in their inputs outside of the $Q$, $K$, and $V$ matrices, which all of them receive. The COO and CSR algorithms have no additional attention-specific parameters, $P_a$, but their sparse mask (graph), $G$, is an explicit input. The local, 1D dilation, 2D dilation, and global approaches have attention-specific parameters, $P_a$, as described above, but no explicit mask is provided. The ``\texttt{\textbf{Get\_Neighbors}}" function will, depending on the inputs, calculate the mask indices (neighbors) that will be operated on for each row. Then, for a given row, $i$, the algorithm pulls in neighbor information for $K$ and $V$ to calculate a subset of the attention output, adjusting for changes in the softmax statistics with each new neighbor that is pulled from.

The algorithms proposed are truly sparse, meaning that for any arbitrary mask provided, they will conduct the attention operation with the minimum number of calculations required. This means that, as the mask becomes more sparse, the amount of work required decreases. 

\noindent \textbf{Work Optimality:} The serial complexity of computing attention with a mask is $O(S_f \times L^2 \times d)$ assuming $d$ as the same dimension of $K$, $Q$, and $V$. Assuming a CRCW PRAM model~\cite{akl1999power}, state-of-the-art implementations that require a dense-dense matrix multiplication followed by invalidation of zero entries, followed by a sparse matrix multiplication using $p$ processors will have a cost of $O(L^2d + S_fL^2d)$ which is not cost optimal (the cost of a parallel algorithm should be equal to the serial complexity for cost/work optimality). In contrast, our algorithm only performs computations for the non-zero elements of the mask and achieves a cost of $O(S_f \times L^2 \times d)$, making it work optimal.  

\section{Experimental Results}
\label{sec:ex_results}

\subsection{Algorithm Programming and Verification}
\label{subsec:alg_prog_ver}

Each of the custom graph processing algorithms discussed in this paper were written in NVIDIA's CUDA programming language which is designed to be run on NVIDIA GPUs. The CUDA kernels were written with and compiled by PyTorch's custom extension system which required the Ninja build tool. This process created a binding that exposed the functions so that they could be called directly from Python and interact with PyTorch tensor objects. These PyTorch back-ends of the algorithms will be open-sourced. The software versions utilized are:

\begin{itemize}
    \item \textbf{Python:} 3.10.8
    \item \textbf{PyTorch:} 2.4.0 (CUDA 12.1)
    \item \textbf{CUDA:} 12.1.1
    \item \textbf{Ninja:} 1.11.1
\end{itemize}

In order to verify correctness, all variants of the algorithms written were tested against PyTorch's \texttt{scaled\_dot\_product\_attention} function that utilized the C++ back-end as it allows for arbitrary binary masks to be passed in. The verification process entailed creating a mask as a PyTorch tensor and converting it into the desired sparse matrix representation, or making sure that the mask matched the implicit one that would be utilized by the ordered sparsity algorithms. The query, key, and value matrices had context lengths of 256 and embedded dimensions of 32; each was created from the uniform random distribution $[0, 1)$ and was identical for both functions. Resulting outputs were compared using PyTorch's \texttt{allclose} function with an absolute tolerance of $1 \times 10^{-08}$, a relative tolerance of $1 \times 10^{-05}$, and \texttt{NaN} values set to equal. The outputs were deemed identical for attention with varied levels of sparsity. 

\subsection{System Specifications}
\label{subsec:sys_spec}

All empirical results were captured using three different HPC systems. Within this paper, each system will be referred to by its associated GPU. The specifications of each cluster are shown below in Table \ref{tab:hpc_specs}.

\begin{table}[!ht]
\centering
\caption{The specifications of all three HPC systems utilized to gather empirical data for this paper.}
\label{tab:hpc_specs}
    \begin{tabular}{ |M{1.8cm}|M{1.3cm}|M{2.0cm}|M{1.8cm}| } 
        \hline
        \textbf{GPU} & \textbf{GPU Driver} & \textbf{CPU} & \textbf{Operating System} \\ 
        \hline
        NVIDIA A100 (SXM4 80GB) & 555.42.02 & AMD EPYC 7742 & Red Hat Linux 8.10 \\ 
        \hline
        NVIDIA L40 (48GB) & 560.35.03 & Intel Xeon Gold 5418Y & Red Hat Linux 8.7 \\ 
        \hline
        NVIDIA V100 (SXM2 32GB) & 560.35.03 & Intel Xeon Gold 6226 & Red Hat Linux 8.7 \\ 
        \hline
    \end{tabular}
\end{table}

For all empirical experiments, each system was configured using the Slurm workload manager to have one GPU, eight cores, and 32GB of memory. All software was identical between systems and matched that mentioned in Section \ref{subsec:alg_prog_ver}.

\subsection{Microbenchmarking}
\label{subsec:microbench}

\subsubsection{\textbf{Objective:} To analyze the comparative runtime performance of the graph processing algorithms across varied systems and problem dimensions.}

\subsubsection{Setup and Results}

A series of microbenchmarks were conducted across all three systems shown in Table \ref{tab:hpc_specs}. All six graph processing algorithms were measured as well as PyTorch's SDP attention that utilized a mask. Each algorithm, except for the COO variant, was run with context lengths ($L$) of 8,192; 16,384; and 24,576, embedded dimensions ($d_k$) of 64; 128; and 256, and sparsity factors ($S_f$) in the range $(0, 1]$. The COO implementation was only run with a context length of 8,192, all three embedded dimensions, and sparsity factors in the range $(0, 0.4]$ due to its long runtime. A dilation factor of 1 was used for both the 1D dilation and 2D dilation masks. The local, 1D dilation, and 2D dilation masks calculated window/block size to fit the associated sparsity factor. Additionally, the NVIDIA V100 GPU does not have data for a context length of 24,576 due to memory restrictions. Each combination of input parameters were run 10 times for a warm up and then an additional 15 iterations were timed for the benchmark. 

\begin{figure*}[!htbp]
     \centering
        \begin{subfigure}[b]{\textwidth}
         \centering
         \includegraphics[width=\textwidth]{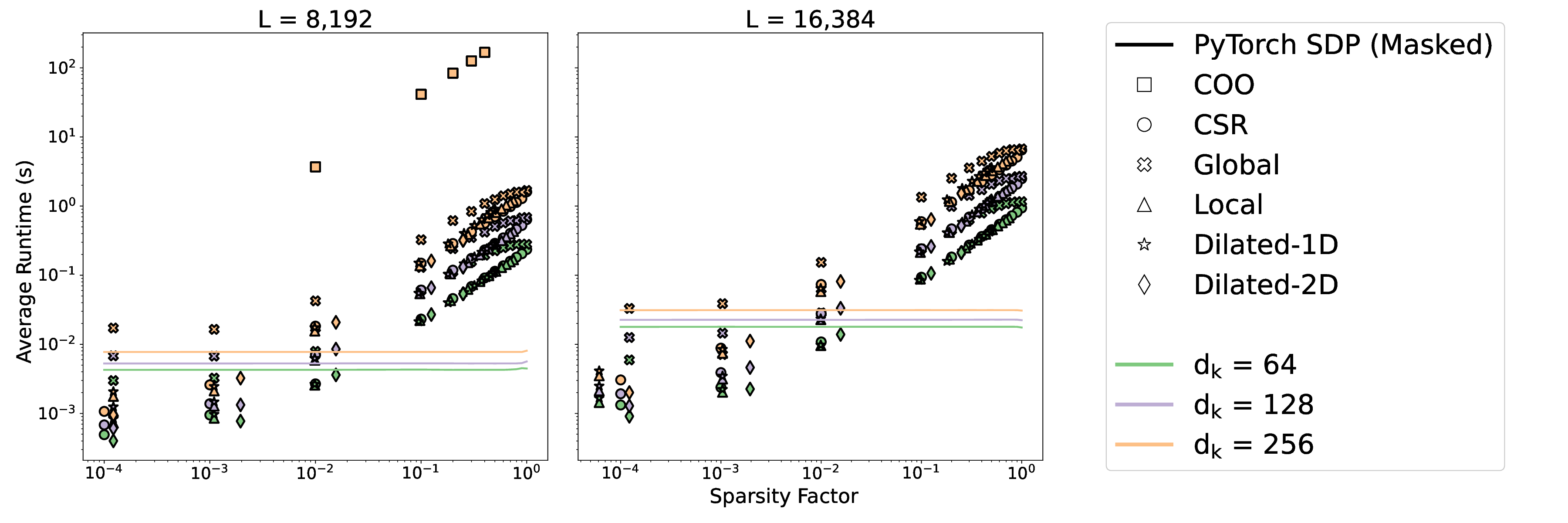}
         \caption{\textbf{NVIDIA V100}}
         \label{fig:ex_1_v100}
     \end{subfigure}
     \hfill
     \begin{subfigure}[b]{\textwidth}
         \centering
         \includegraphics[width=\textwidth]{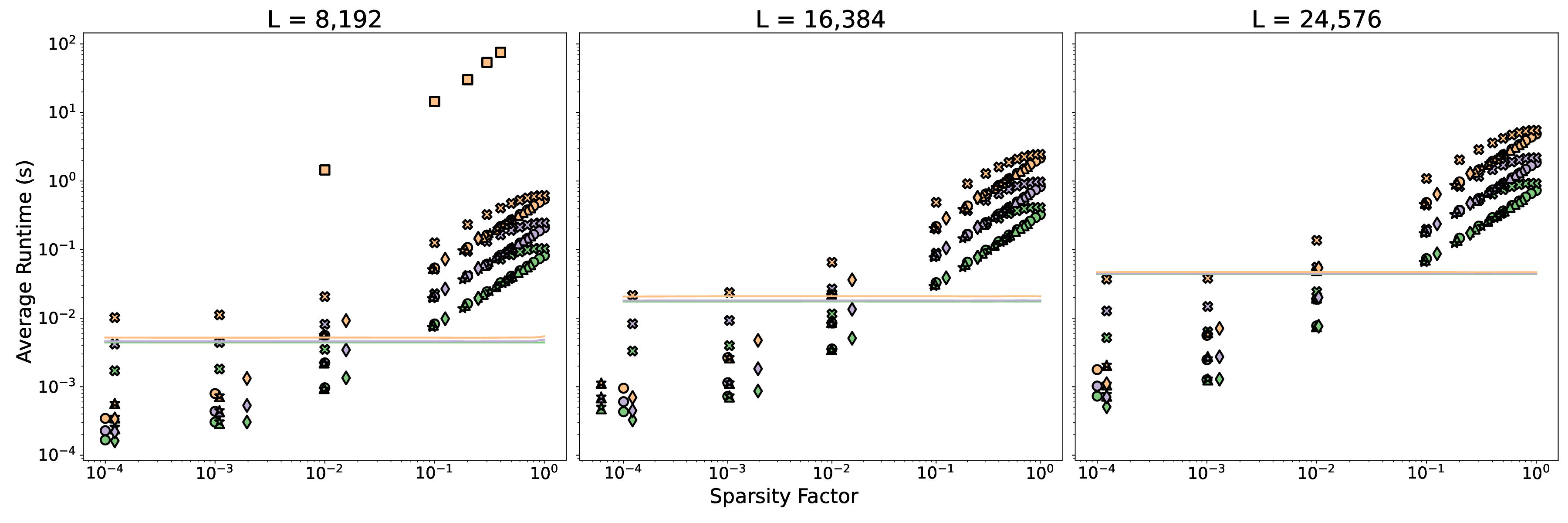}
         \caption{\textbf{NVIDIA L40}}
         \label{fig:ex_1_l40}
     \end{subfigure}
     \hfill
     \begin{subfigure}[b]{\textwidth}
         \centering
         \includegraphics[width=\textwidth]{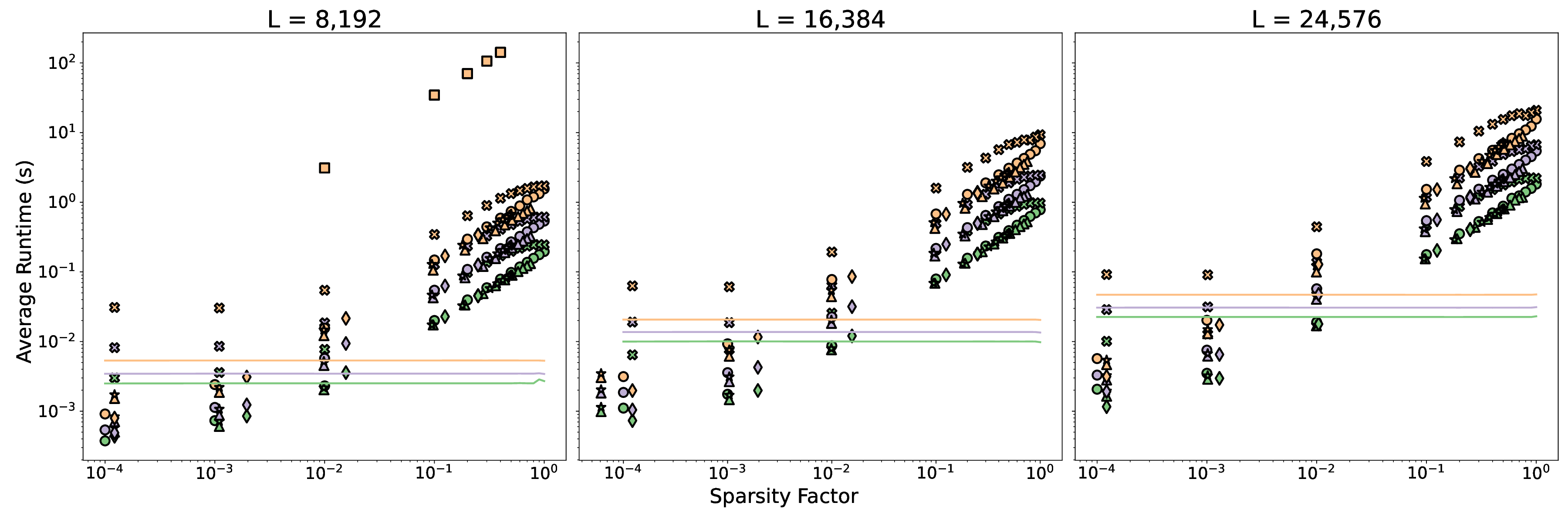}
         \caption{\textbf{NVIDIA A100}}
         \label{fig:ex_1_a100}
     \end{subfigure}
        \caption{Plotting the average log runtime performance from 15 benchmark runs of our algorithms and PyTorch's C++ back-end SDP attention implementation across three NVIDIA GPUs as the context length ($L$), embedded dimension ($d_k$), and sparsity factor ($S_f$) vary. $L$ varies between each plot, increasing from left-to-right in a row (corresponding to a decrease in sparsity). $d_k$ varies by color within a plot. Log $S_f$ increases within a plot along the x-axis. \textbf{(a)} shows the runtimes for the NVIDIA V100 system, \textbf{(b)} has the runtimes for the NVIDIA L40 system, and \textbf{(c)} presents the runtimes for the NVIDIA A100 system.}
        \label{fig:ex_1_results}
\end{figure*}

The benchmark results are plotted in Figure \ref{fig:ex_1_results} where \ref{fig:ex_1_v100} corresponds to all data from the NVIDIA V100, \ref{fig:ex_1_l40} the NVIDIA L40, and \ref{fig:ex_1_a100} the NVIDIA A100. The y-axis is the log-scale average runtime in seconds while the x-axis is the log sparsity factor. Within a row of graphs, as one moves from left to right, the context length increases. As shown in the key in \ref{fig:ex_1_v100}, the colors correspond to different embedded dimensions while the different symbols are the algorithms. The PyTorch SDP implementation is portrayed as a line rather than a symbol. 

% \FloatBarrier

% \FloatBarrier

\subsubsection{Analysis}

Within Figure \ref{fig:ex_1_results}, there is a noticeable uniformity of the trends between both different GPUs and $L$. Across all algorithms, except for COO, one can easily observe an increase in average runtime as $d_k$ increases in size. This is expected as growth along $d_k$ corresponds directly to the number of multiplications and additions required for each dot product operation. For COO, all the markers are close to one another. Due to COO being a selection of ordered coordinates (grouped rows and sorted columns), the current algorithm must search to find the limits of a row before running the attention process; the search cost grows as the algorithm strays farther from row zero. The incurred cost is unique to COO as none of the other attention methods must search to find bounds before they can run. The ordered sparsity approaches do calculate indices, which is similar, however the operation cost does not increase with later tokens in the sequence.

The SDP algorithm demonstrates constant performance with increases to $S_f$. This is due to the fact that the function conducts a fully dense matrix multiplication before masking and then conducts another dense matrix multiplication regardless of the sparsity of the attention matrix. SDP outperforms the graph processing algorithms until greater sparsity is achieved ($S_f \leq 0.01$). With greater sparsity, fewer operations are conducted by the graph processing algorithms and, outside of the COO form, they are able to complete the operation faster than the PyTorch function. Outside of shared memory utilization, the current forms of the written algorithms are naive and untuned, with future optimizations the performance crossover with SDP should occur at a lower sparsity. Further optimizations are discussed in Section \ref{subsec:lim}.

Amongst the graph processing algorithms, the local, 1D dilation, 2D dilation, and CSR implementations are relatively similar in performance. They demonstrate the best average runtimes across all GPUs, $L$, $d_k$, and levels of $S_f$. Their trends are also identical due to their similar implementations. The global variant has comparable performance to CSR at high $S_f$, but demonstrates a slower decrease in runtime with increased sparsity. This is due to the form of sparsity present in a global mask: certain rows are entirely masked while others are entirely non-masked (minus the identity diagonal as the smallest local size was chosen for benchmarking, and global token columns). Because parallelization occurs along the $L$ dimension, certain rows will be fully dense (minus one) as long as $S_f$ is not 1. This can create an imbalanced distribution of work amongst CUDA blocks, and the algorithm can only be as fast as its slowest block. This disparity is unlikely to happen in the CSR implementation and will not happen in the local variant for higher levels of sparsity.

2D dilation had the best average performance across all $d_k$ and $L$ for $S_f$ $< 0.001$, showing speedups over SDP of 13.37x, 42.12x, and 11.88x for the V100, L40, and A100, respectively. With the same conditions, 1D dilation demonstrated speedups over SDP of 6.74x, 26.40x, and 6.95x for the V100, L40, and A100, respectively. Local observed average speedups over SDP of 7.87x, 27.56x, and 8.07x for the V100, L40, and A100 respectively. The global algorithm, with the same context, saw average speedups over SDP of 1.40x, 2.87x, and 0.87x for the V100, L40, and A100 respectively. CSR showed the best average speedup for explicit masks with speedups over SDP of 9.85x, 31.59x, and 7.81x for the V100, L40, and A100, respectively. With $S_f < 0.1$, COO saw speedups over SDP of 0.002x, 0.003x, and 0.001x on the V100, L40, and A100, respectively. The search within the kernel was harmful to COO's performance. Moving forward, empirical results will focus on CSR for explicit mask exploration.

Across all of the GPUs, the L40 demonstrated the fastest average runtimes, possibly because it was the newest GPU tested. The L40's crossover in performance with the CSR and local algorithms appears to happen at a higher sparsity factor, closer to 0.1, than the other GPUs. Additionally, its SDP runtimes are much closer together than the other algorithms. In the future, we will profile each of the algorithms across all GPUs to gain a better understanding of trade-offs between GPUs.

\subsection{Theoretical Context Length Limits}

\subsubsection{\textbf{Objective:} To quantitatively understand, using various algorithms, limitations on context length introduced by the amount of memory an accelerator has.}

\subsubsection{Setup and Results}

\begin{figure*}[!ht]
     \centering
          \begin{subfigure}[b]{\textwidth}
         \centering
         \includegraphics[width=0.9\textwidth]{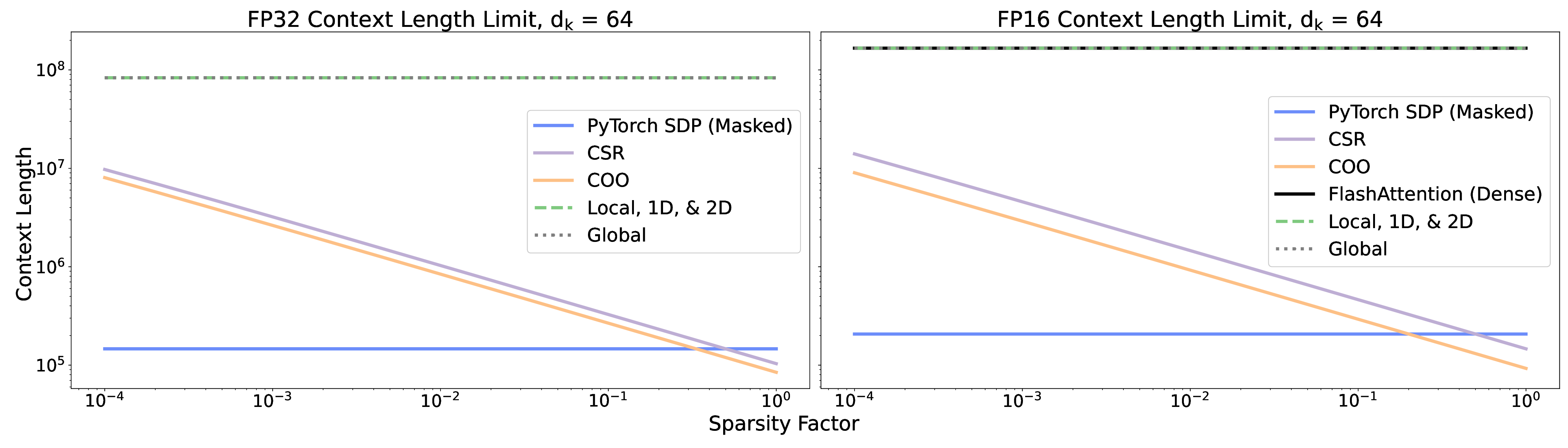}
         \vspace{-0.2cm}
         \caption{}
         \label{fig:ex_3_d64}
     \end{subfigure}
     % \hfill
     \begin{subfigure}[b]{\textwidth}
         \centering
         \includegraphics[width=0.9\textwidth]{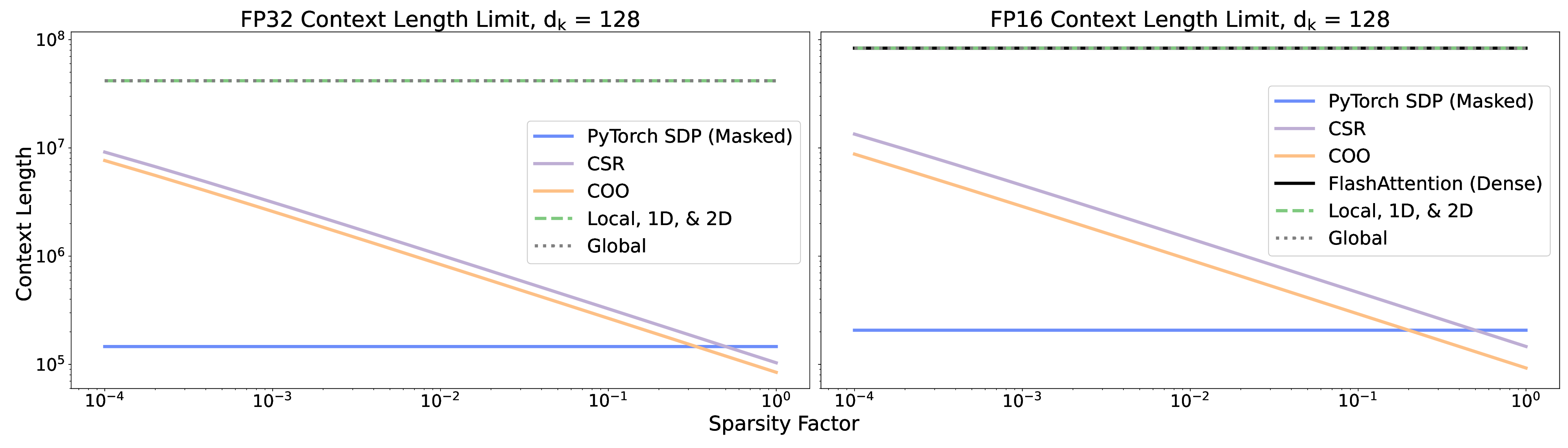}
         \vspace{-0.2cm}
         \caption{}
         \label{fig:ex_3_d128}
     \end{subfigure}
        \caption{Plots showing the trend in maximum log context length ($L$) achievable by different algorithms as the log sparsity factor ($S_f$) increases along the x-axis. This is for single-headed attention on one NVIDIA A100 GPU. The left plots showcase performance for 32-bit floating-point values and the right for 16-bit floating-point values. The embedded dimension changes between \textbf{(a)} and \textbf{(b)}. \textbf{(a)} shows $L$ for $d_k = 64$ and \textbf{(b)} shows $L$ for $d_k = 128$.}
        \label{fig:ex_3_results}
\end{figure*}

The theoretical context length limits for a given sparsity were calculated by solving inequalities that relate the total GPU memory to the amount of memory occupied by tensors during runtime. The NVIDIA A100 GPU, which has an 80 GB capacity, was considered for the calculations. The results are show in Figure \ref{fig:ex_3_results} where \ref{fig:ex_3_d64} and \ref{fig:ex_3_d128} showcase theoretical log context length limits for embedded dimensions of 64 and 128, respectively, as the log sparsity factor increases along the x-axis. All plots portray performance of a single-batch, single-headed attention implementation. The left-most graphs relate to 32-bit floating-point (FP32) tensors while the right-most relate to 16-bit floating-point (FP16) tensors. FlashAttention does not operate on FP32 data. Table \ref{tab:max_l_trend} tabulates the data presented in Figure \ref{fig:ex_3_results}. It also shows the theoretical limit of all attention mechanisms on one A100 using dimensions from the Llama 3 series 8 billion parameter model: 32 heads and $d_k$ of 4,096 \cite{dubey2024llama3herdmodels}.

% \FloatBarrier

\begin{table*}[!ht]
\centering
\caption{The theoretical maximum achievable context lengths ($L$) for high-sparsity attention with one NVIDIA A100 GPU for select attention algorithms. The following are given: the data type, sparsity factor ($S_f$), and embedded dimension ($d_k$).}
\label{tab:max_l_trend}
    \begin{tabular}{ |M{0.5cm}|M{0.7cm}|M{0.6cm}|M{0.7cm}|M{1.1cm}|M{1.1cm}|M{1.1cm}|M{1.8cm}|M{1.3cm}|M{1.3cm}|M{1.3cm}| M{1.3cm}|} 
        \hline
        \textbf{Data Type} & \textbf{$S_f$} & \textbf{$d_k$} & \textbf{Heads} & \textbf{Max $L$ - SDP (Masked)} & \textbf{Max $L$ - CSR} & \textbf{Max $L$ - COO} & \textbf{Max $L$ - FlashAttention (Dense)} & \textbf{Max $L$ - Local} & \textbf{Max $L$ - Global} & \textbf{Max $L$ - Dilated (1D)} & \textbf{Max $L$ - Dilated (2D)} \\ 
        \hline
        \multirow{3}{2em}{FP32} & 0.0001 & 64 & 1 & 146,416 & 9,732,519 & 8,038,418 & \textit{Unsupported} & \textbf{83,235,801} & 83,235,769 & \textbf{83,235,801} & \textbf{83,235,801} \\ 
        \cline{2-12}
        & 0.0001 & 128 & 1 & 146,288 & 9,152,140 & 7,644,258 & \textit{Unsupported} & \textbf{41,779,838} & 41,779,830 &  \textbf{41,779,838} &  \textbf{41,779,838} \\ 
        \cline{2-12}
        & 0.0001 & 4,096 & 32 & 25,651 & 950,434 & 865,272 & \textit{Unsupported} & \textbf{1,305,620} & \textbf{1,305,620} & \textbf{1,305,620} & \textbf{1,305,620} \\ 
        \hline
        \multirow{3}{2em}{FP16} & 0.0001 & 64 & 1 & 207,116 & 14,013,926 & 9,009,893 & \textbf{166,471,601}* & \textbf{166,471,601} & 166,471,472 & \textbf{166,471,601} & \textbf{166,471,601} \\ 
        \cline{2-12}
        & 0.0001 & 128 & 1 & 206,988 & 13,416,404 & 8,764,655 & \textbf{83,559,676}* & \textbf{83,559,676} & 83,559,643 & \textbf{83,559,676} & \textbf{83,559,676} \\ 
        \cline{2-12}
        & 0.0001 & 4,096 & 32 & 36,381 & 1,601,190 & 1,200,336 & \textbf{2,611,240}* & \textbf{2,611,240} & 2,611,239 & \textbf{2,611,240} & \textbf{2,611,240} \\ 
        \hline
    \end{tabular}
    \smallskip
    \parbox[t]{\textwidth}{\footnotesize
        \textit{\hspace{0.3cm}*:}
        It should be noted that FlashAttention is a dense operation and it does not account for the sparsity.
    }
\end{table*}

% \FloatBarrier

Empirical tests were run on the NVIDIA A100 system from Table \ref{tab:hpc_specs} with the software described in Section \ref{subsec:alg_prog_ver} and node setup from Section \ref{subsec:sys_spec}. Three algorithms were benchmarked at set context lengths: FlashAttention, local, and CSR. The local implementation had $S_f$ values (except for 160 million) according to the relation from Section \ref{ssec:how_sparse}, CSR does not for $L$ over 1.6 million due to memory restrictions. Every test utilized the FP16 data type. For each configuration, an algorithm had ten warm up runs and then was timed for 15 benchmark runs with the average runtime reported in Table \ref{tab:max_l_run}. FlashAttention, for a context length of 160 million, was the only exception, there was no warm up and only one benchmark run due to the large amount of time required for a single iteration. 

% For a context length of 16 million, CSR could not utilize a sparsity factor of $1 \times 10^{-4}$ due to memory restrictions, so $4 \times 10^{-5}$ was used instead; the local algorithm showcases results for both sparsity factors.

% \begin{table}[!ht]
% \centering
% \caption{The average runtimes of FlashAttention, the local attention implementation, and the CSR version on a series of long context lengths ($L$).}
% \label{tab:max_l_run}
%     \begin{tabular}{ |M{1.6cm}|M{2.0cm}|M{1.6cm}|M{1.6cm}| } 
%         \hline
%         \textbf{$L$} & \textbf{Algorithm} & \textbf{Sparsity Factor} & \textbf{Average Runtime (s)} \\ 
%         \hline
%         \multirow{2}{5em}{160,000,000} & FlashAttention & --- & 37,477.25 \\ 
%         \cline{2-4}
%         & Local & 0.00001 & \textbf{733.93} \\ 
%         \hline
%         \multirow{3}{4.5em}{16,000,000} & FlashAttention & --- & 372.35 \\ 
%         \cline{2-4}
%         & Local & 0.00004 & \textbf{29.28} \\
%         \cline{2-4}
%         & CSR & 0.00004 & 32.46 \\
%         \hline
%         \multirow{3}{4em}{8,000,000} & FlashAttention & --- & 92.88 \\ 
%         \cline{2-4}
%         & Local & 0.0001 & \textbf{18.62} \\
%         \cline{2-4}
%         & CSR & 0.0001 & 20.49 \\
%         \hline
%         \multirow{3}{4em}{1,600,000} & FlashAttention & --- & 3.48 \\ 
%         \cline{2-4}
%         & Local & 0.0001 & \textbf{0.81} \\
%         \cline{2-4}
%         & CSR & 0.0001 & 0.86 \\
%         \hline
%     \end{tabular}
% \end{table}

\begin{table}[!ht]
\centering
\caption{The average runtimes of FlashAttention, the local attention implementation, and the CSR version on a series of long context lengths ($L$).}
\label{tab:max_l_run}
    \begin{tabular}{ |M{1.6cm}|M{2.0cm}|M{1.6cm}|M{1.6cm}| } 
        \hline
        \textbf{$L$} & \textbf{Algorithm} & \textbf{Sparsity Factor} & \textbf{Average Runtime (s)} \\ 
        \hline
        \multirow{2}{5em}{160,000,000} & FlashAttention & --- & 37,477.25 \\ 
        \cline{2-4}
        & Local & 0.00001 & \textbf{733.93} \\ 
        \hline
        \multirow{3}{4.5em}{16,000,000} & FlashAttention & --- & 372.35 \\ 
        \cline{2-4}
        & Local & 0.00017 & \textbf{124.67} \\
        \cline{2-4}
        & CSR & 0.00004 & 32.46* \\
        \hline
        \multirow{3}{4em}{8,000,000} & FlashAttention & --- & 92.88 \\ 
        \cline{2-4}
        & Local & 0.00034 & \textbf{62.32} \\
        \cline{2-4}
        & CSR & 0.0001 & 20.49* \\
        \hline
        \multirow{3}{4em}{1,600,000} & FlashAttention & --- & \textbf{3.48} \\ 
        \cline{2-4}
        & Local & 0.0017 & 12.46 \\
        \cline{2-4}
        & CSR & 0.0017 & 13.67 \\
        \hline
    \end{tabular}
    \parbox[t]{\textwidth}{\footnotesize
        \textit{\hspace{0.1cm}*:}
        CSR did not use the same sparsity according to Section \ref{ssec:how_sparse} due to \\
        \hspace*{0.3cm} memory restrictions.
    }
\end{table}

\subsubsection{Analysis}

As demonstrated in Figure \ref{fig:ex_3_results} and Table \ref{tab:max_l_trend}, greater sparsity in the attention matrix allows the CSR and COO implementations to theoretically perform attention with context lengths nearly two orders of magnitude longer than an SDP implementation. The CSR variant possesses a longer context length than COO because of its sparse matrix structure: CSR has one vector with memory complexity of $\mathcal{O}(L)$ and two others with $\mathcal{O}(S_fL^2)$ while COO has all three vectors with $\mathcal{O}(S_fL^2)$ memory complexity. FlashAttention, local, 1D dilation, and 2D dilation algorithms can all achieve even greater context lengths independent of the sparsity because they utilize an online softmax and do not require explicit storage of the attention matrix, only two statistics vectors, each with memory complexity of $\mathcal{O}(L)$ \cite{dao2022flashattentionfastmemoryefficientexact, milakov2018onlinenormalizercalculationsoftmax}. Global is similar, except it has one additional input to indicate which tokens require global attention with maximum length $L$. As shown in Table \ref{tab:max_l_trend}, this minutely reduces its maximum context length in relation to the other implicit mask algorithms.

Table \ref{tab:max_l_run} shows that sparse operations can outperform hardware-optimized algorithms like FlashAttention for extremely long context lengths. For $L$ of 1.6 million, local and CSR were 0.28x and 0.25x as fast as FlashAttention, respectively. However, as $L$ increases and $S_f$ is adjusted according to Section \ref{ssec:how_sparse}, local saw speedups of 1.49x and 2.99x over FlashAttention for $L$ of 8 million and 16 million, respectively. For $L$ of 160 million, we observed a speedup of 51.06x over FlashAttention. This trend is expected, as less work is being conducted by the GPU. Between sparse implementations, local outperformed CSR for $L$ of 1.6 million because it does not have to read from and write to global memory for the attention matrix.

\subsection{Runtime Trade-Off Given a Context Length}

\subsubsection{\textbf{Objective:} To observe the runtime effects of sparsity on a graph processing algorithm in comparison to the dense FlashAttention.}

\subsubsection{Setup and Results}

All benchmarks were run on the NVIDIA A100 system from Table \ref{tab:hpc_specs}. Two algorithms were used: FlashAttention and our local implementation. The FP16 data type was used for compatibility with FlashAttention. The local window size (length a token can see behind or ahead) was held constant in one set of tests for decreasing sparsity with increasing context length and was dynamic in the other to achieve a constant sparsity. FlashAttention does not take sparsity into account as it is a dense operation. For each variable combination, a given algorithm had ten warm up runs and was timed for 15 runs afterwards with the average runtime reported. 

\begin{figure*}[!htbp]
    \centering
    \includegraphics[width=\textwidth]{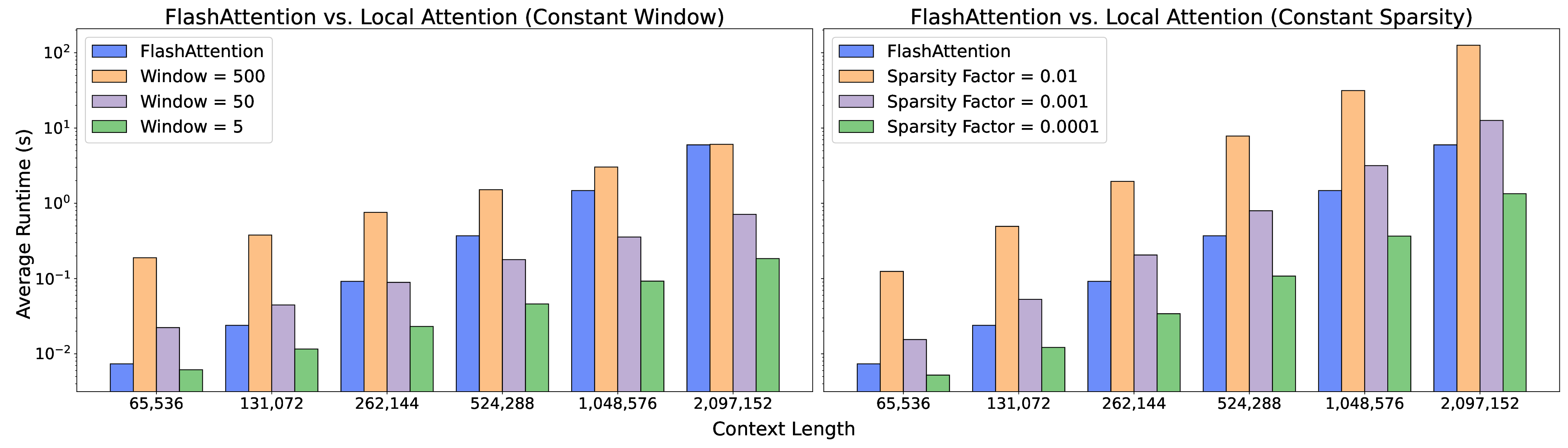}
    \vspace{-0.4cm}
    \caption{Plotting the average log runtime performance from 15 benchmark runs of PyTorch's FlashAttention implementation against our local attention with either constant window size (left plot) or constant sparsity (right plot). The x-axis shows increasing context length from left-to-right.}
    \label{fig:ex_2_a100}
\end{figure*}

In Figure \ref{fig:ex_2_a100} the left chart shows the average log runtime with a constant window size whereas the right shows a constant sparsity in relation to the context length. 

% \FloatBarrier

% \FloatBarrier

\subsubsection{Analysis}

In the left-most plot of Figure \ref{fig:ex_2_a100}, one can see that increasing sparsity in the local algorithm does allow it to outperform FlashAttention. However, with a fixed low sparsity, the gap in performance continues to grow with increased context length. This can be seen in Table \ref{tab:max_l_run} and the right plot of Figure \ref{fig:ex_2_a100}, where with a low sparsity factor of 0.0001 and $L$ of 65,536 the local variant saw a 1.41x speedup over FlashAttention that increased to a 4.46x improvement for $L$ of 2,097,152.

\subsection{Performance on Popular Attention Masks}

\subsubsection{\textbf{Objective:} To explore the runtime of our algorithms in relation to PyTorch's SDPA while using standard attention masks.}

\subsubsection{Setup and Results}

All benchmarks were collected from the NVIDIA A100 system from Table \ref{tab:hpc_specs}. PyTorch's SDP served as a baseline comparison for the graph processing algorithms as FlashAttention does not support arbitrary masking. The sparsity patterns utilized are shown in Figure \ref{fig:ex_4_masks}, where the left mask was modelled in two ways: a combination of our local and global algorithm run sequentially and then only our CSR approach. The central mask was run only with our CSR algorithm. The right-most mask was constructed in two forms: a sequential combination of our local, global, and CSR (for random attention) algorithms and then only our CSR variant. For all approaches the local size was set to 50 in each direction and three global tokens were used, the dilated local window used a dilation factor of two giving an effective local size of 100, and $S_f = 0.001$ for the random sparsity in the third mask. Each masking pattern was tested at four different context lengths where each algorithm had ten warm up runs and 15 timed runs where the average was reported. For a given context length, each approach was given identical masks. The outputs of each approach were deemed identical in a manner similar to that described in Section \ref{subsec:alg_prog_ver}. 

\begin{figure*}[!ht]
    \centering
    \includegraphics[width=\textwidth]{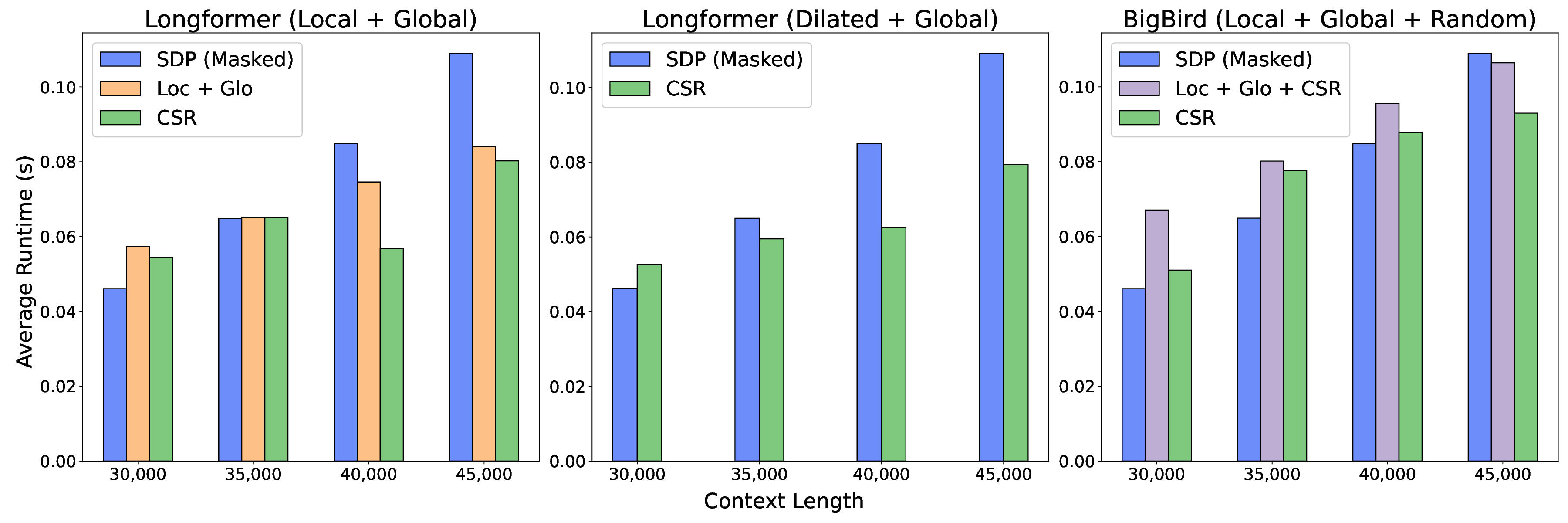}
    \vspace{-0.4cm}
    \caption{Plotting the average runtime performance from 15 benchmark runs of different attention mechanisms from well-known masked transformer models as the context length increases along the x-axis. The left-most plot is Longformer with local and global attention; it was modelled with PyTorch's SDP, a double kernel call of our local and global, and CSR. The middle plot is Longformer with dilated local and global attention; it was run with PyTorch's SDP and our CSR function. The right-most plot is BigBird with local, global, and random attention; it was run with PyTorch's SDP, a sequential kernel call of our local; global; and CSR functions, and just our CSR implementation.}
    \label{fig:ex_4_a100}
\end{figure*}

Figure \ref{fig:ex_4_a100} showcases the average runtime performance for each algorithm on each mask type as the context length increases along the x-axis.

% \FloatBarrier

% \FloatBarrier

\subsubsection{Analysis}

Within Figure \ref{fig:ex_4_a100} one can observe that for all attention mask implementations the SDP function has identical average runtimes for set context lengths, confirming the lack of sparse operations. Additionally, as the context length increases, the performance of the graph processing approaches improves relative to SDP. This result aligns with the observation from other experiments that increased sparsity benefits the graph processing algorithms when compared to dense methods. 

For the two Longformer implementations, both sparse results match or outperform the SDP implementation at a context length of 35,000. This swap in performance is not observed in BigBird until a context length of 45,000 and is likely due to the addition of a third masking operation that reduces the sparsity of the mask. Within the sparse methods, there is a consistent trend that a single call to the CSR implementation performs as well as or better than sequential calls to functions that apply a specific form of attention.

\section{Discussion}

\subsection{Limitations and Future Directions}
\label{subsec:lim}

Our current implementations only support a single head and, with trivial extensions, can support multi-head attention. This drives our immediate next step. Another limitation of our work is that the graph algorithm implementation is naive and does not use GPU specific optimizations such as better utilization of shared cache, data layout optimization to avoid bank conflicts, or using tensor cores (systolic arrays) to bolster performance. We will incorporate these optimizations in our future work.

We would also like to note that the benefits of utilizing graph processing algorithms are realized for very sparse attention masks ($S_f < 0.01$). As shown in LongNet~\cite{ding2023longnet}, one can still obtain high accuracy with exponentially decreasing sparsity for extremely long context lengths. But for shorter context lengths used in Natural Language Tasks (a few thousand to tens of thousands), graph processing algorithms may not perform as well as the state-of-the-art dense versions such as FlashAttention. 

Furthermore, as alternatives to custom-written algorithms, we would also like to explore the representation of our algorithms using performant functions from graph processing libraries like GraphBLAS and cuSPARSE \cite{GraphBLAS7}. Moreover, to support distributed training across multiple nodes, we will implement distributed memory versions of the algorithms discussed in this paper along with graph partitioning techniques to load balance work across the nodes. Finally, we also plan to explore more sophisticated sparse matrix representation formats for specific attention mask patterns to reduce their storage overheads. 

\subsection{Impact}
\label{subsec:impact}

Using a single NVIDIA A100 GPU with 80 GB memory, we demonstrated that our graph algorithm based attention implementations could compute attention for a context length of 160 million at a fraction of the time of FlashAttention --- the most efficient attention mechanism implementation that exists currently. In a training workflow, even if we assume that only $25\%$ of memory is available for attention computations (with the remaining being used for gradients and other overheads), only 32 GPUs will be needed to reach a context length of 1 billion that is required for high-impact scientific applications such as genomics~\cite{nguyen2024hyenadna}.

\section{Conclusion}

We created six graph processing algorithms with open-source PyTorch back-ends that perform the sparse (masked) attention operation and benchmarked their performance in terms of runtime and achievable context length for comparison to other dense and sparse attention approaches. Coupled with high sparsity, the introduction of graph processing techniques allowed our algorithms to achieve greater context lengths and lower runtimes than other masked approaches. Our ordered sparse approaches achieved identical context lengths to the state-of-the-art FlashAttention algorithm (dense attention) and were able to outperform it in same-context length situations when sparsity was introduced. The utilization of sparse attention accompanying other graph processing techniques makes extremely long context lengths achievable. 

\section*{Acknowledgements}
This research was supported by NSF Awards 2425535, 2411447, and 2117439. All work made use of the High Performance Computing Resource in the Core Facility for Advanced Research Computing at Case Western Reserve University. 

\bibliographystyle{IEEEtran}
\IEEEtriggeratref{50}
\bibliography{ref}

\end{document}